\RequirePackage{amsmath}
\PassOptionsToPackage{table}{xcolor} 
\documentclass[runningheads,a4paper]{llncs}
\usepackage[caption=false]{subfig}
\usepackage[capitalise]{cleveref}
\usepackage{amssymb}
\usepackage{graphicx}
\usepackage{wrapfig}
\usepackage{xcolor}
\usepackage{color}
\usepackage{tikz}
\usepackage{url}
\usepackage{multirow}
\usepackage{gensymb}
\usepackage{booktabs}
\usepackage{tabularx}
\usepackage{array}
\usepackage{float}

\newcommand{\Mypm}{\mathbin{\tikz [x=1.4ex,y=1.4ex,line width=.1ex] \draw (0.0,0) -- (1.0,0) (0.5,0.08) -- (0.5,0.92) (0.0,0.5) -- (1.0,0.5);}}%
\newcommand{\todo}[1]{}
\renewcommand{\todo}[1]{{\color{red} TODO: {#1}}}

\urldef{\mailsa}\path|{****|
\urldef{\mailsb}\path|****|
\urldef{\mailsc}\path|****|

\usepackage{etoolbox}
\usepackage{color,soul} 
\definecolor{yellow}{rgb}{1.0,1.0,0.5}
\definecolor{green}{rgb}{0.5,1.0,0.5}

\setlength\belowcaptionskip{-15pt}

\graphicspath{{./images/}}

\makeatletter
\let\origsection\section
\renewcommand\section{\@ifstar{\starsection}{\nostarsection}}

\newcommand\nostarsection[1]
{\sectionprelude\origsection{#1}\sectionpostlude}

\newcommand\starsection[1]
{\sectionprelude\origsection*{#1}\sectionpostlude}

\newcommand\sectionprelude{%
  \vspace{-2mm}
}

\newcommand\sectionpostlude{%
  \vspace{0mm}
}
\makeatother

\begin{document}

\mainmatter  

\title{Context-Sensitive Super-Resolution for Fast Fetal Magnetic Resonance Imaging}

\titlerunning{Super-Resolution for Under-Sampled MRI}

\author{Steven McDonagh$^{1}$ \and Benjamin Hou$^{1}$ \and Amir Alansary$^1$ \and Ozan Oktay$^1$ \and Konstantinos Kamnitsas$^1$ \and Mary Rutherford$^2$ \and Jo V. Hajnal$^2$ \and Bernhard Kainz$^{1,2}$}
\authorrunning{S. McDonagh et al.}

\institute{
	$^1$Biomedical Image Analysis Group, Imperial College London \\
	$^2$Division of Imaging Sciences and Biomedical Engineering, King’s College London
}

%

\maketitle

\begin{abstract}
\label{sec:abstract}
3D Magnetic Resonance Imaging (MRI) is often a trade-off between fast but low-resolution image acquisition and highly detailed but slow image acquisition. Fast imaging is required for targets that move to avoid motion artefacts. This is in particular difficult for fetal MRI. Spatially independent upsampling techniques, which are the state-of-the-art to address this problem, are error prone and disregard contextual information.
In this paper we propose a context-sensitive upsampling method based on a residual convolutional neural network model that learns organ specific appearance and adopts semantically to input data allowing for the generation of high resolution images with sharp edges and fine scale detail. 
By making contextual decisions about appearance and shape, present in different parts of an image, we gain a maximum of structural detail at a similar contrast as provided by high-resolution data.
We experiment on $145$ fetal scans and show that our approach yields an increased PSNR of $1.25$ $dB$ when applied to under-sampled fetal data \emph{cf.} baseline upsampling. Furthermore, our method yields an increased PSNR of $1.73$ $dB$ when utilizing under-sampled fetal data to perform brain volume reconstruction on motion corrupted captured data. 
\end{abstract}

\vspace{-2mm}
\section{Introduction}
\vspace{-1mm}
\label{sec:intro}
\parskip0pt
Currently, 3D imaging of moving objects is limited by the time it takes to acquire a single image. The slower an imaging modality is, the more likely motion induced artefacts will occur within and between individual slices of a 3D volume. 
Very fast imaging modalities like Computed Tomography are not always applicable because of harmful ionising radiation, and ultrasound often suffers from poor image quality. Thus, Magnetic Resonance Imaging (MRI) is usually the modality of choice when; large fields of view, high anatomical detail, and non-invasive imaging is required. MRI is often applied to image involuntary moving objects such as the beating heart and examination of the fetus in-utero. Motion compensation for cardiac imagining can be achieved through ECG gating.
However, fetal targets do not provide options for gated or tracked image acquisition to compensate for motion. Thus motion compensation is performed during post-processing of oversampled input spaces, usually involving the acquisition of orthogonally oriented stacks of slices~\cite{kainz2015fast}. Oversampling with high resolution (HR) slices causes long scan times, which is uncomfortable and risky for patients like pregnant women. This limits the possible number of scan sequences during examination. However, improving image resolution is key to improving accuracy, understanding of anatomy and assessment of organ size and morphology. Imaging at lower resolution increases acquisition speed, thus partly mitigating the likelihood for motion between individual slices but at the cost of missing structural detail that could render the scan inappropriate for diagnostic purposes. Due to signal-to-noise ratio (SNR) limitations, the acquired slices are usually also \emph{thick} compared to the in-plane resolution and thus negatively influence the visualization of anatomy in 3D.

Na\"ive up-sampling of fast but low resolution (LR) images is undesirable for the clinical practice, since results lack information. Information content cannot be increased by simply increasing the number of pixels with linear interpolation methods. Therefore, optimization-based super-resolution (SR) methods have been explored to generate rich volumetric information from oversampled input spaces. However, these methods are highly dependant on the quality and amount of input samples and depend on the choice of the objective function. Recent work, \emph{e.g.}~\cite{Dong2016a}, on example-based SR has focused on incorporating additional prior image knowledge, and, in particular, deep neural networks have been employed to solve the single-image SR (SISR) problem. However, the majority of recent contributions typically place strong emphasis on natural images and therefore lack domain specific high-frequency detail prior knowledge~\cite{Borman1998}.
\newline
\newline
\noindent \textbf{Contribution:}
We present a novel approach to SISR in the context of motion compensation when using fast to acquire, low resolution volumes. Taking inspiration from recent investigation of network based SR for MRI modalities~\cite{Oktay2016}, we propose a network architecture with convolutional and transposed-convolutional layers and hypothesize that such a deep network architecture can be tailored to context sensitive applications, such as motion compensation of the fetal brain, and yield volume reconstruction improvements from low resolution input. Our network learns subject specific details from potentially motion corrupted input data and accurately reintroduces the expected fidelity allowing motion compensation and high quality reconstruction from fast low resolution input. 

Our model is in particular data-adaptive since the upsampling is performed by learnable transposed-convolution layers instead of a fixed kernel. By performing the upsampling in the final layers of the network we avoid early redundant computation in a HR space, enabling a computational saving. Additionally by considering entire LR in-plane slice samples at training time, in comparison to image \emph{patches}, we gain a large receptive field to enable the learning of spatial context, organ structure and anatomy.  

We evaluate our method on 145 healthy fetal scans. The proposed approach shows improved qualitative results when compared visually to linear methods. Quantitative reconstruction performance, peak signal-to-noise-ratio (PSNR) and structural similarity index measure (SSIM) improve, accordingly. In particular, we reach comparable reconstruction quality with half as many data samples, thus half of the currently required scan time, when compared to motion compensated reconstruction from high-resolution image acquisition.
\newline
\newline
\noindent \textbf{Related work:}
\label{sec:rel_work}
%
The topic of SR has received much attention in the literature and a large body of work exists however, historically, algorithms exhibiting good performance in 2D domains such as satellite or facial imagery, are not necessarily ideal for 3D medical imaging. This is partly due to domain specific effects such as loss of spatial information caused by motion during slow target acquisition. Various algorithms have been shown to produce leading results~\cite{Nasrollahi2014} in differing domains. 


SISR accounts for missing image information by using previously observed examples to optimise the LR-HR mapping between images or patches. In the medical imaging domain, data-adaptive patch-based approaches to SISR reconstruction~\cite{Jia2016single,Manjon2010non} have been shown to prevent the occurrence of well-known blurring effects, often found when using classical interpolation approaches. Interpolation techniques tend to increase the smoothness of images in an isotropic manner, however data-adaptive non-local methods allow for highly anisotropic reconstruction where required. In patch-based methods, the radius of 3D patch used to compute the similarity among voxels is often a free parameter and the choice of receptive field size typically affects computational cost when using iterative optimisation. 

Learning based approaches also allow data-adaptive reconstruction and CNNs in particular have recently been successfully applied to context sensitive SISR for cardiac imaging. The work of \cite{Oktay2016} use a regression architecture based on~\cite{Dong2016a} with a modified $l_1$ objective function. The approach performed SR in the slice-select direction of lowest MRI resolution, \emph{i.e.}, one-dimensionally and utilized transposed convolutional layers at the start of the network architecture to perform the upsampling, \emph{prior} to convolutions, thus learning high level features in latter layers on (spatially) large feature maps. 

Two-dimensional SR is a popular research area in natural image processing due to many applications requiring enhancement of a visual experience while limiting the amount of raw data that needs to be recorded, transferred or stored. Recent network-based approaches such as SRGAN~\cite{Ledig2016} 
apply Generative Adversarial Networks (GAN) 
to achieve large up-sampling factors of up to four. 




Motion compensation for MRI volume reconstruction typically incorporates a SR component. However to the best of our knowledge state-of-the-art network based SR techniques, capable of learning problem and sensor specifics from available data have not been harnessed for the upsampling step found in Slice-to-Volume frameworks for the reconstruction task. In this work we investigate the accuracy advantages that such an approach can contribute to the example of fetal MRI volume reconstruction. 

Contemporary SR components in MRI Slice-to-Volume reconstruction (SVR) tasks perform optimisation based incremental updates to the HR volume estimate. To achieve this, the SR problem for volume reconstruction has been modelled directly by considering minimisation of an error norm function and use of Huber function statistics~\cite{gholipour2010robust} or gradient-weighted averaging~\cite{kim2010intersection}. The ill-posed nature of modelling upsampling requires that the objective be regularised. Gholipour~\emph{et al.}~\cite{gholipour2010robust} add a Tikhonov term to the cost for this purpose while Rousseau~\emph{et al.}~\cite{Rousseau2010,rousseau2006registration,Rousseau2013} select a regularisation term that includes an approximation of Total Variation (TV) to better preserve edges. Tourbier~\emph{et al.}~\cite{tourbier2015efficient} apply fast convex optimization techniques for the SR problem also using an edge-preserving TV regularization. Murgasova~\emph{et al.}~\cite{Kuklisova-Murgasova2012} used intensity matching and complete outlier removal for reconstruction. SR volume intensities are iteratively updated using the error gradients resulting from differences between simulated and observed slice samples. Transforming observed slice information to the upsampled volume space requires accurate yet potentially computationally expensive estimation of the sensor point spread function (PSF) and~\cite{kainz2015fast} developed a fast multi-GPU accelerated implementation for the task. 


\vspace{-2mm}
\section{Method} 
\begin{figure}[!htbp]
\centering
  \includegraphics[width=1.0\linewidth]{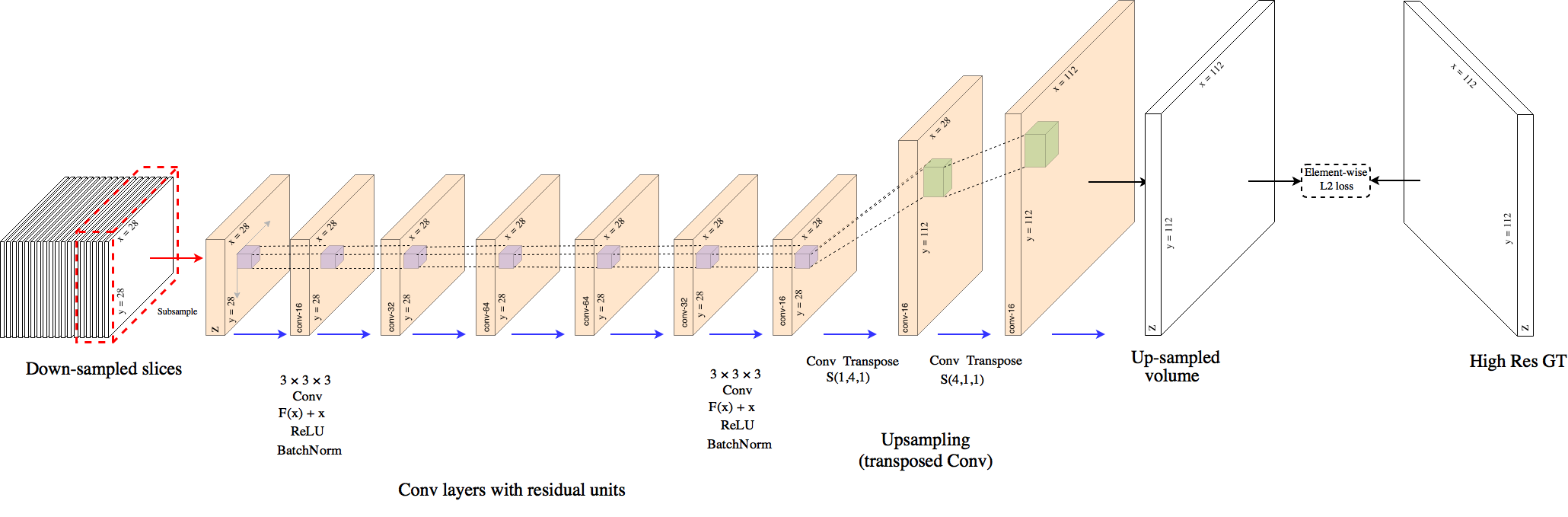}
  \caption{Our proposed CNN network architecture for MRI super-resolution. See text for architecture details.}
  \label{fig:network}
\end{figure}

\begin{figure}[!htbp]
\centering
  \includegraphics[width=1.0\linewidth]{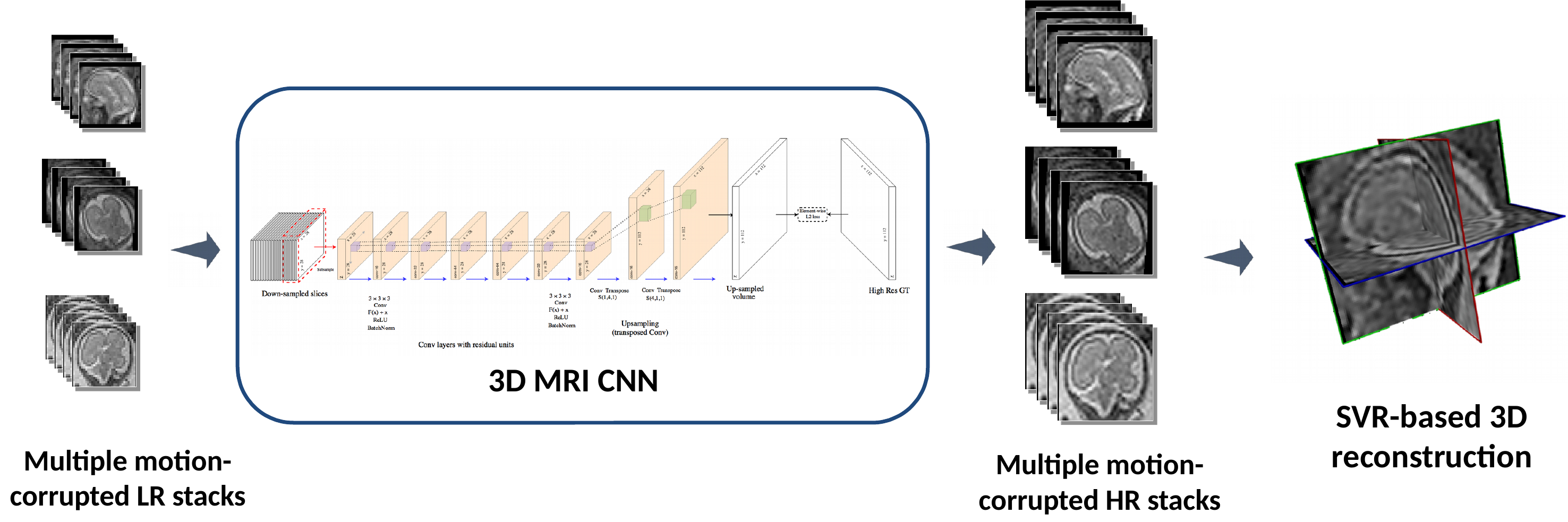}
  \caption{The proposed framework for providing upsampled, high resolution input for motion correction and volume reconstruction.}
  \label{fig:pipeline}
\end{figure}

\vspace{-1mm}
\label{sec:method}
The proposed approach implements a fully three-dimensional CNN architecture to infer upsampled MRI imagery, enabling HR input to be provided for subsequent SVR and motion compensation tasks. We define an architecture utilising 3D volumetric convolutions that have recently been shown to add value for medical imaging tasks considering 3D imagery~\cite{Kamnitsas2016deepmedic,Cciccek20163d}. Fig.~\ref{fig:network} provides a schematic of our upsampling network and architecture design details are provide in the \textbf{3D MRI CNN} subsection below. Fig.~\ref{fig:pipeline} provides a schematic diagram indicating where the upsampling network component is implemented in a SVR reconstruction framework.


The architecture differs from recent network based MRI SR models~\cite{Oktay2016} by generating feature maps in the LR image space \emph{cf.} early redundant feature channel upsampling or fixed kernels~\cite{Dong2015}, reducing memory and computation requirements while retaining the flexibility of learnable upsampling layers. As previously reported~\cite{Shi2016}, early upsampling tends to introduce redundant computation in the HR space since no additional information is added into the model by performing transposed convolutions at an early stage of the architecture.

\noindent Our approach mitigates the acquisition quality cost of low resolution imagery by considering the problem of estimating a high dimensional $\mathbf{y} \in \mathbb{R}^M$, for a given observation $\mathbf{x} = f(y) \in \mathbb{R}^N$ where $(N << M)$. SR is an underdetermined inverse problem, and as such the function $f$ performs a downsampling and is typically non-invertible. The low-dimensional observation $\mathbf{x}$ is mapped to the high-dimensional $\mathbf{y}$ by recovery through the MR image acquisition model~\cite{Greenspan2009super}, a series of operators such that: $\mathbf{x}=DBSM\mathbf{y}+\eta$ where $M$ defines a spatial displacement, \emph{e.g.} due to motion, $S$ is the slice selection operator, $B$ is the point-spread function (PSF) used to blur the selected slice, $D$ is a decimation operator, and $\eta$ is a Rician noise model. We approximate solutions to this inverse problem by estimating $\phi(\mathbf{x},\Theta)$ from the LR input such that a cost, defined between $\phi(\mathbf{x},\Theta)$ and $\mathbf{y}$, is minimized. We estimate the parameters $\Theta$ using a CNN architecture with parameters $\Theta$ that parametrise network layers to model the distribution $p(y|x)$. Training samples are defined as ($\mathbf{x}_i,\mathbf{y}_i$).
\newline
\newline
\noindent \textbf{3D MRI CNN:}
In-plane, low-resolution MRI stacks are synthetically generated simply by filtering HR images with a Cosine Windowed Sinc blurring kernel followed by a decimation operator to provide LR-HR training pairs as input. Training samples consist of entire LR in-plane imagery with a volume defined by $z>=1$ out-of-plane slices forming 3D volume training samples, providing contextual information from multiple slices. Here we report on experimental upsampling factors of $\times2$,$\times4$ and $z=5$.


Our 3D-CNN architecture contains nine layers consisting of six convolutional layers, utilising standard ReLU activations and residual units, followed by two transposed-convolutional layers (with corresponding strides of two or four) and a final single-channel layer to build the full resolution output. 
The ReLU activation function has exhibited strong performance when upscaling both natural images~\cite{Dong2016a} and MRI 3D volume data~\cite{Oktay2016}. Intermediate feature maps $h^{(n)}_j$ at layer $n$ are computed through convolutional kernels $w^n_{kj}$ as $max(0 , \Sigma^K_{k=1} h_k^{(n-1)} \ast w^n_{kj}) = h^n_j$ where $\ast$ is the convolutional operator. We follow the common frugal strategy \cite{Simonyan2014} of applying small $(3\times3\times3)$ convolution kernels and spending compute-budget alternatively on layer count to increase receptive field size. 

By introducing two transposed convolution layers we perform the upscaling on in-plane sampling dimensions. In this manner, upscaling weights are learned specifically for the SR task where $(x \uparrow U_x) * w_j = h_j^0$ and $(h_1 \uparrow U_y) * w_j = h_j^1$ where $\uparrow$ is a zero-padding upscaling operator and $\{U_x,U_y\} = M/N$ are the in-plane upscaling factors. This allows for explicit optimization of the upsampling filters and facilitates training in an end-to-end manner for the SR task. By implementing trainable upsampling layers we improve upon the alternative strategy of initial independent linear upsampling, followed only by convolutional layers, as we gain an ability to learn upsampling weights specific to the SR task. In practice this often improves MRI image signal quality in image regions close to boundaries~\cite{Oktay2016}. Residuals learned by the convolution layers and the upscaled transposed-convolutional output are used to reconstruct the final HR image. This allows the regression function to learn non-linearities such as the high frequency components of the signal.

Training involves evaluating the error function $\Psi_{l_2}(\cdot)$ that calculates the difference between the reconstructed HR images and the ground truth volumes that were down-sampled to provide training data. Model weights are updated using standard back-propagation and adaptive moment estimation. 
In comparison to modified $l_1$ losses~\cite{Oktay2016} or recent \emph{perceptual-quality} SR objective functions~\cite{Ledig2016}, we implement a standard voxel-wise $l_2$ loss function to provide gradient information and emphasize voxel-wise difference to the ground-truth. An implementation of our model training strategy is made available online\footnote{https://github.com/DLTK}.
\newline

\noindent \textbf{Fetal Brain Volume Reconstruction:} We combine our SR network with Slice-to-Volume registration (SVR) \cite{kainz2015fast}. SVR requires multiple orthogonal stacks of 2D slices to provide improved reconstruction quality. By upsampling stacks prior to reconstruction we provide a means to acquire larger sets of low-resolution input. 
The motion-free 3D image is then reconstructed from the upsampled slices and motion-corrupted and misaligned areas are excluded during the reconstruction using an EM-based outliers rejection model.

\vspace{-2mm}
\section{Experiments}
\vspace{-1mm}
\label{sec:experiments}
\noindent\textbf{Data:} 
We test our approach on clinical MR scans with varying gestational age. All scans have been ethically approved. The dataset contains $145$ MR scans of healthy fetuses at gestational age between $20$--$25$ weeks. The data has been acquired on a Philips Achieva $1.5$T, the mother lying $20\degree$ tilt on the left side to avoid pressure on the inferior vena cava. Single-shot fast spin (ssFSE) echo T2-weighted sequences are used to acquire stacks of images that are aligned to the main axes of the fetus. Three to six stacks with a voxel size of $1.25mm \times 1.25mm \times 2.5mm$  per stack are acquired for the whole womb. Imagery is manually masked and cropped to isolate fetal brain regions. 
\newline
\newline
\noindent\textbf{Experimental details:} We employ our 3D MRI network and separately two baseline SR strategies to upsample image stack inputs that serve as input to the SVR pipeline. SVR then performs motion compensation and volume reconstruction. We assess upsampled image quality directly and, additionally, investigate the effect of the proposed upsampling strategy on reconstruction quality, from the (initially) low resolution fetal data. We report three quantitative metrics: PSNR, structural similarity index  (SSIM) 
and cross-correlation. In the first experiment, the data is randomly split into two subsets and used to train ($100$) and test ($45$) with our SR network. MRI stacks represent $46$ individual patients and all image stacks, belonging to a particular patient, are found uniquely in either the train or test set. Images are cropped, intensity normalised and linearly downsampled by factors of $2$ and $4$ with respect to the in-plane stack axes. This resampling provides LR images to our network resulting in multiple training samples per volume with corresponding ground-truth label (HR source image). The network uses these training pairs to learn the LR to HR mapping. Note that image volume size choices introduce a trade-off between available contextual information and pragmatic memory constraints.


\section{Evaluation and Results}
\label{sec:results}
\begin{figure}[H]
     \centering
     \subfloat{\includegraphics[height=1.7cm]{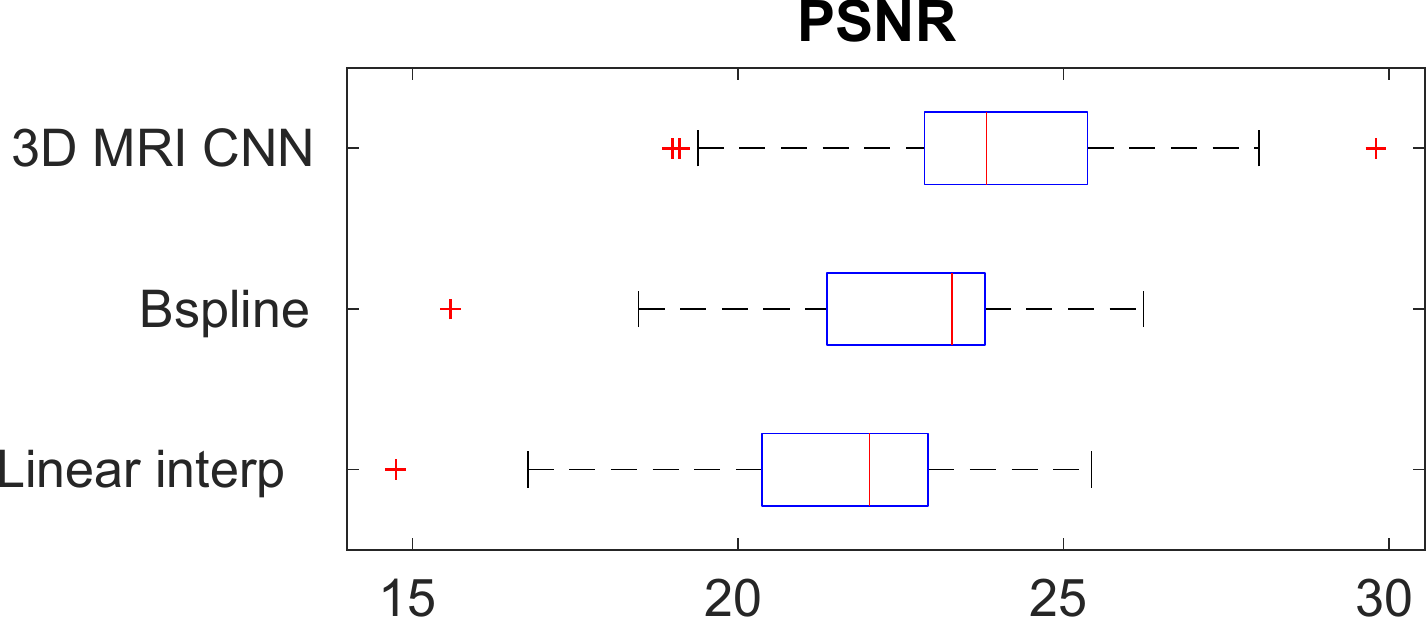}\label{}}\hfill
     \subfloat{\includegraphics[height=1.7cm]{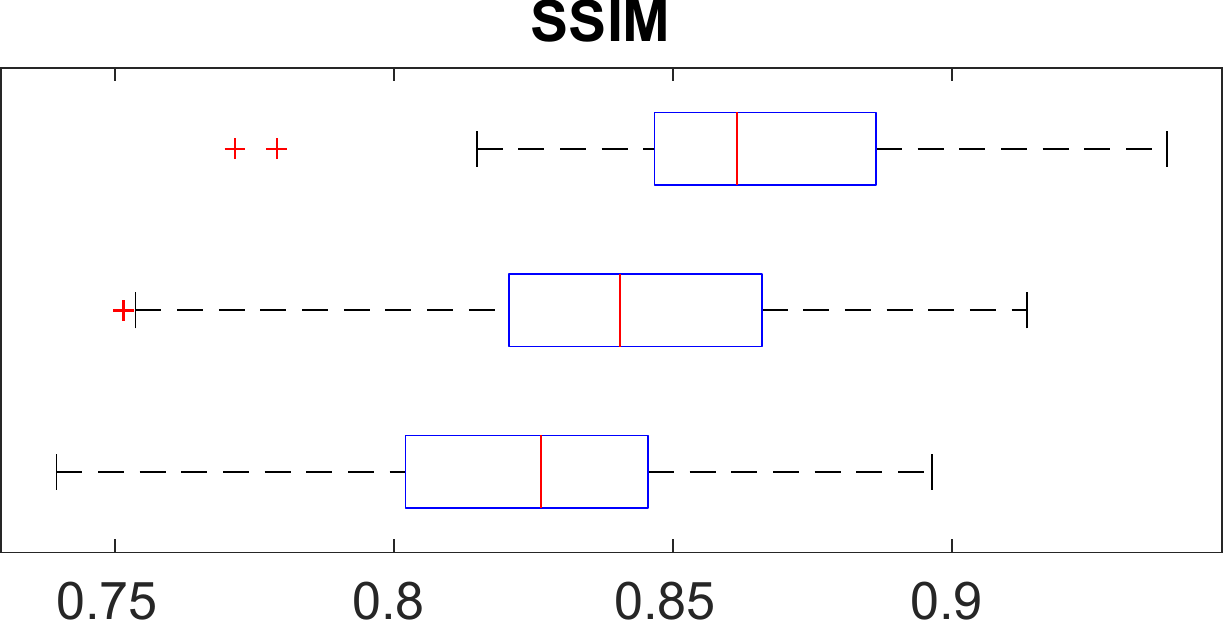}\label{}}\hfill
     \subfloat{\includegraphics[height=1.7cm]{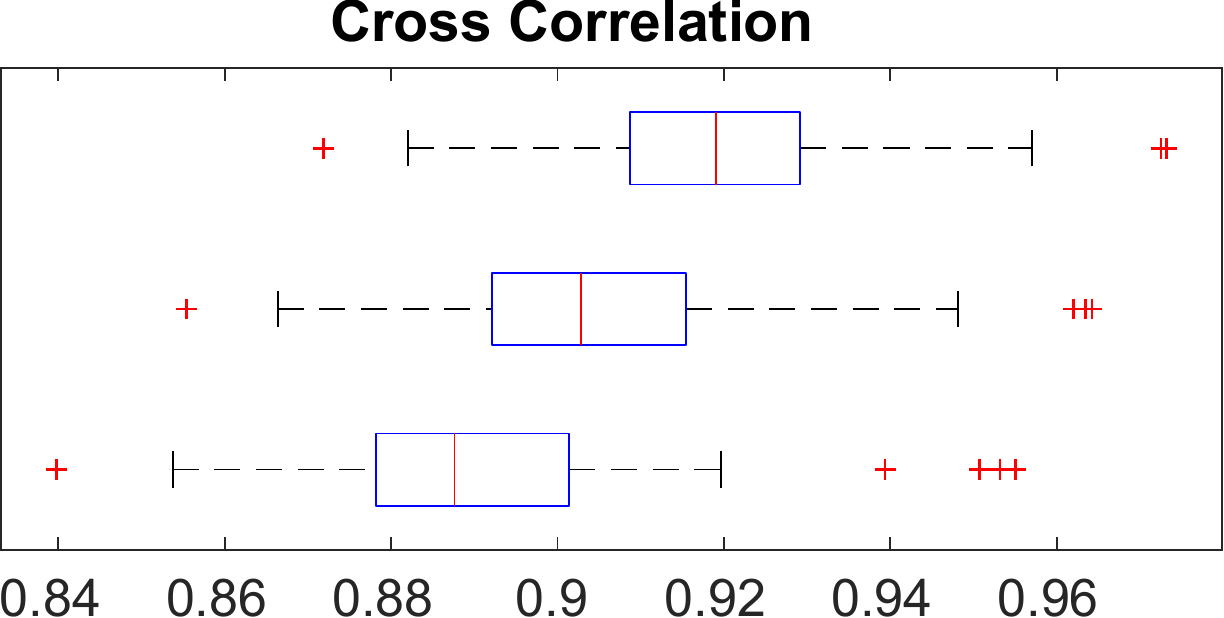}\label{}}
     \caption{PSNR, SSIM and Cross Correlation metrics for $45$ LR image stacks with voxel spacing $(2.50\times2.50\times1.25$)mm that are upsampled $\times2$ in-plane (X,Y) and compared to ground-truth image stacks $(1.25\times 1.25\times1.25)$mm using Linear, B-Spline, 3D MRI CNN methods.}
     \label{fig:boxplot}
\end{figure}

\noindent \textbf{Image Quality Assessments:} We compare HR ground-truth 3D volumes with upsampled LR raw data by measuring PSNR, SSIM and cross-correlation. We report SSIM, in particular, due to the well-understood metric properties that afford assessment of local structure correlation and reduced noise sensitivity. LR test imagery is upsampled in-plane ($X,Y$) by factors of $2$, $4$ to align with target ground-truth resolution. Quality metrics in Fig.~\ref{fig:boxplot} report improvements observed for an image upsampling factor of $2$. This provides initial evidence in support of our hypothesis; \emph{learning problem and sensor specific deconvolutional filters to perform MRI stack upsampling is of benefit for subsequent resolution-sensitive tasks such as motion compensation and HR volume reconstruction}.
\vspace{-9.5mm} 

\begin{figure}[H]
\centering%
\resizebox{0.65\textwidth}{!}{
	\begin{tabular}{>{\centering}m{0.02\textwidth} >{\centering}m{0.15\textwidth} >{\centering}m{0.15\textwidth} >{\centering}m{0.15\textwidth} >{\centering}m{0.15\textwidth} >{\centering\arraybackslash}m{0.15\textwidth}}
		&\textbf{LR input($\times2$)} & \textbf{Linear} & \textbf{B-spline} & \textbf{3D MRI CNN} & \textbf{GT HR} \\
		\rotatebox[origin=c]{90}{\textbf{Axial}}&
            \includegraphics[width=0.15\columnwidth,angle=00,trim= 40 40 40 40,clip]{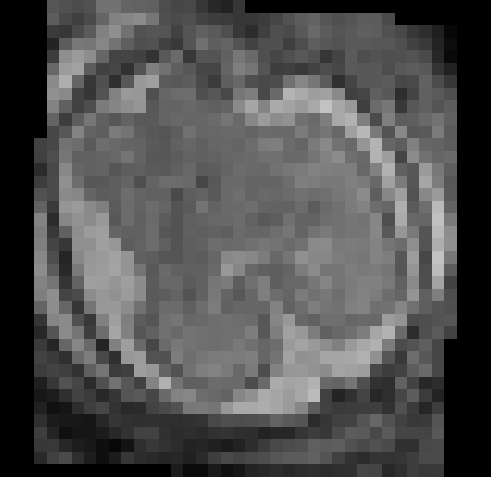} 	&
			\includegraphics[width=0.15\columnwidth,angle=00,trim= 40 40 40 40,clip]{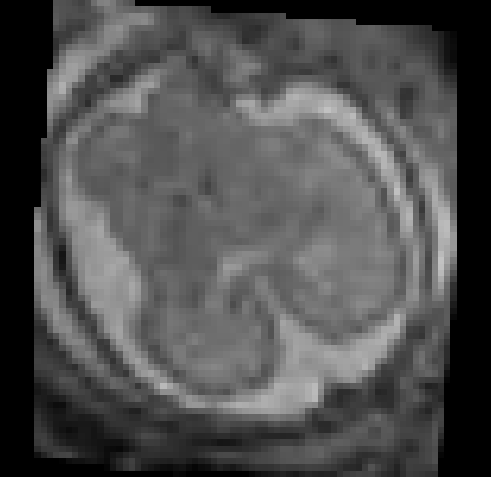}     	&
 			\includegraphics[width=0.15\columnwidth,angle=00,trim= 40 40 40 40,clip]{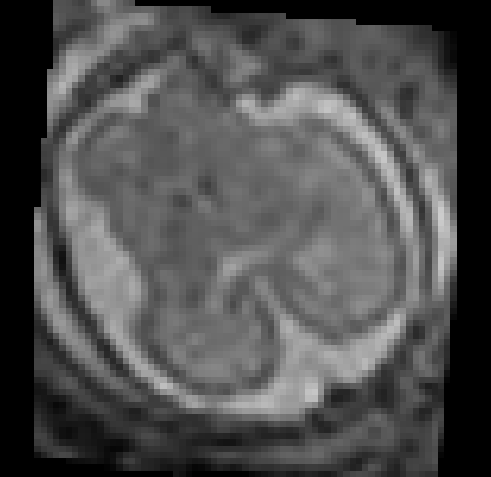}    	&
			\includegraphics[width=0.15\columnwidth,angle=00,trim= 40 40 40 40,clip]{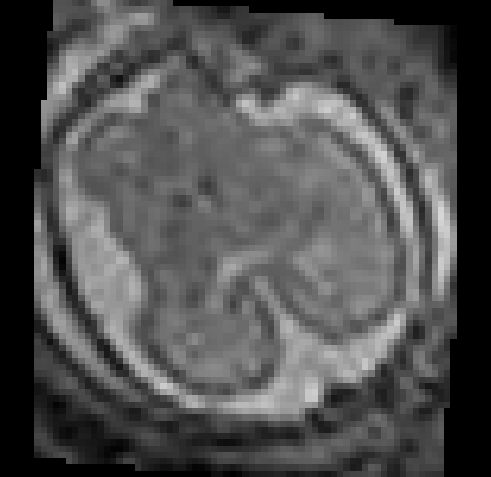}			&
			\includegraphics[width=0.15\columnwidth,angle=00,trim= 40 40 40 40,clip]{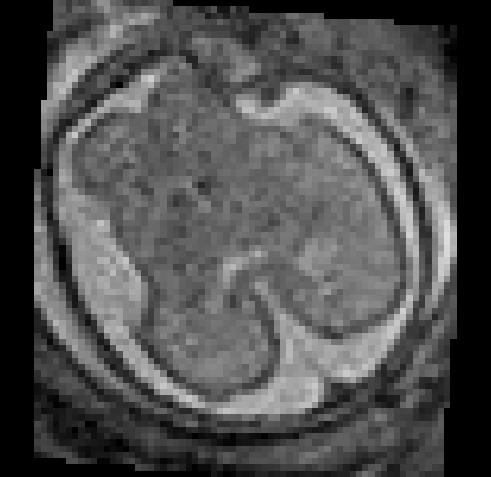}
		\\
		\rotatebox[origin=c]{90}{\textbf{Sagittal}}&
			\includegraphics[height=0.15\columnwidth,angle=90,trim= 40 40 40 40,clip]{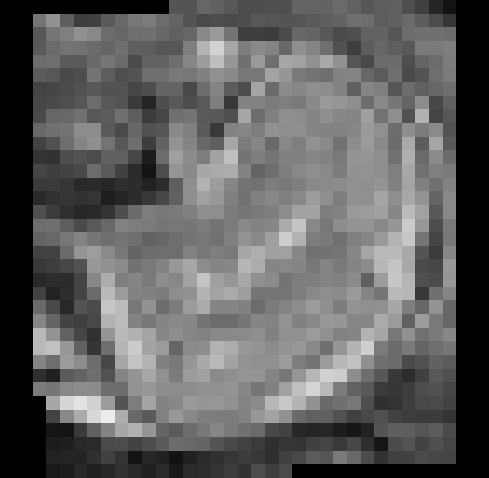} 	&
			\includegraphics[height=0.15\columnwidth,angle=90,trim= 40 40 40 40,clip]{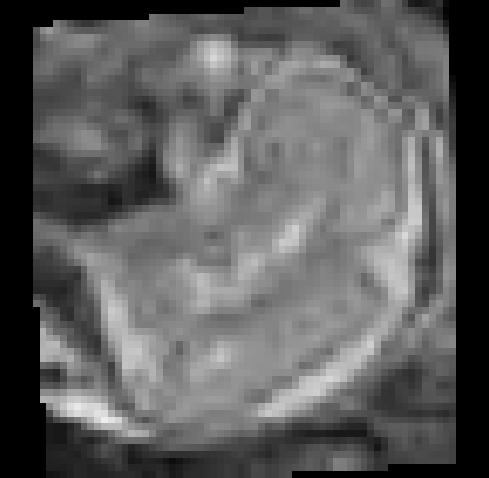}     	&
			\includegraphics[height=0.15\columnwidth,angle=90,trim= 40 40 40 40,clip]{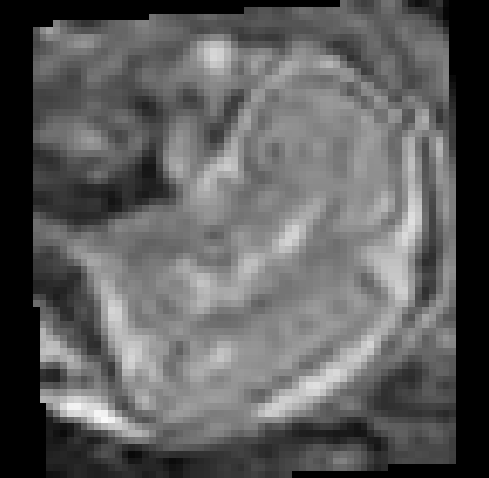}    	&
			\includegraphics[height=0.15\columnwidth,angle=90,trim= 40 40 40 40,clip]{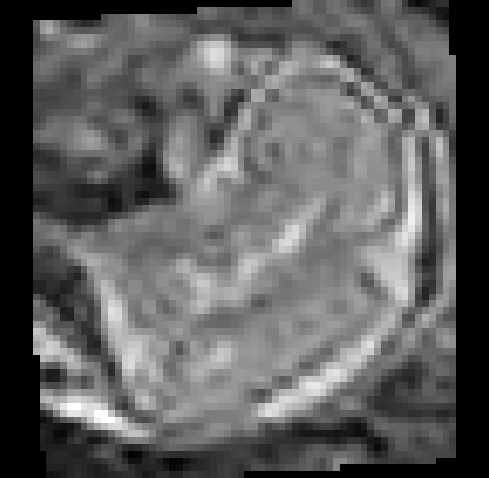}        	&
			\includegraphics[height=0.15\columnwidth,angle=90,trim= 40 40 40 40,clip]{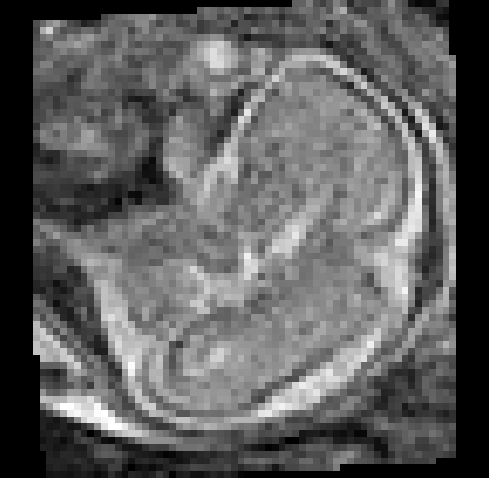}
		\\
		\vspace{-8pt}\rotatebox[origin=c]{90}{\textbf{Coronal}}&
			\vspace{-8pt}\includegraphics[width=0.15\columnwidth,angle=180,trim= 40 40 40 40,clip]{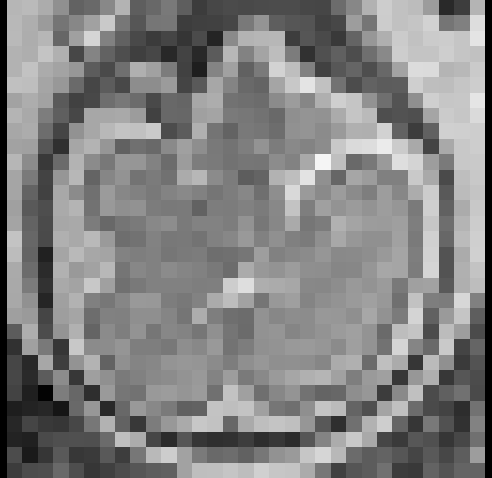}	&
			\vspace{-8pt}\includegraphics[width=0.15\columnwidth,angle=180,trim= 40 40 40 40,clip]{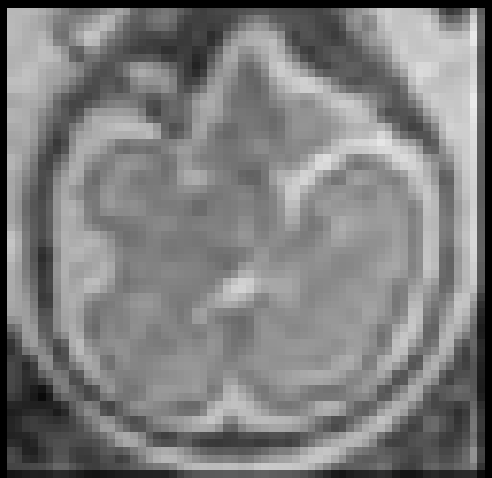}		&
			\vspace{-8pt}\includegraphics[width=0.15\columnwidth,angle=180,trim= 40 40 40 40,clip]{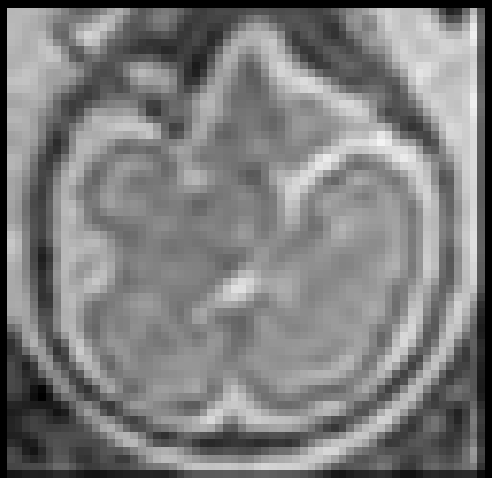}		&
			\vspace{-8pt}\includegraphics[width=0.15\columnwidth,angle=180,trim= 40 40 40 40,clip]{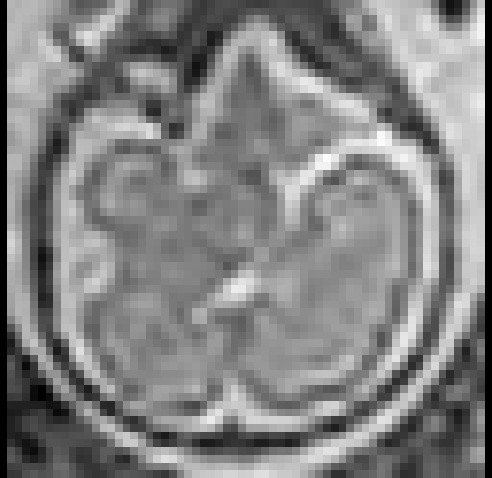}			&
			\vspace{-8pt}\includegraphics[width=0.15\columnwidth,angle=180,trim= 40 40 40 40,clip]{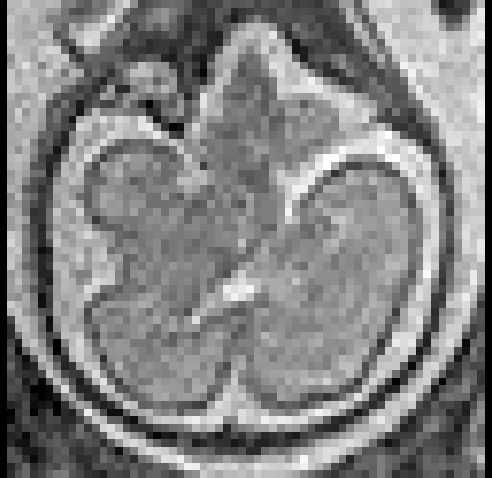} 
		\end{tabular}
}
\caption{Orthogonal fetal MRI stacks showing in-plane stack axes per row. Low resolution input (left) is upsampled by two baselines (col \emph{Linear},\emph{B-spline}) and our learning based approach (col \emph{3D MRI CNN}) \emph{cf.} ground-truth (GT) HR imagery. The learning based 3D MRI CNN, with modality specific priors, provides improved high frequency signal components \emph{cf.} baselines.}
\label{fig:input_upsample}
\end{figure}

\noindent By learning problem specific HR synthesis models, our 3D MRI CNN strategy outperforms the na\"ive baseline up-sampling, quantitatively improving the quality of the inferred HR imagery. Fig.~\ref{fig:input_upsample} exhibits an example of qualitative improvement in orthogonal fetal MRI test-stack axes. 

\begin{figure}[!htbp]
     \centering
     \subfloat{\includegraphics[height=35mm]{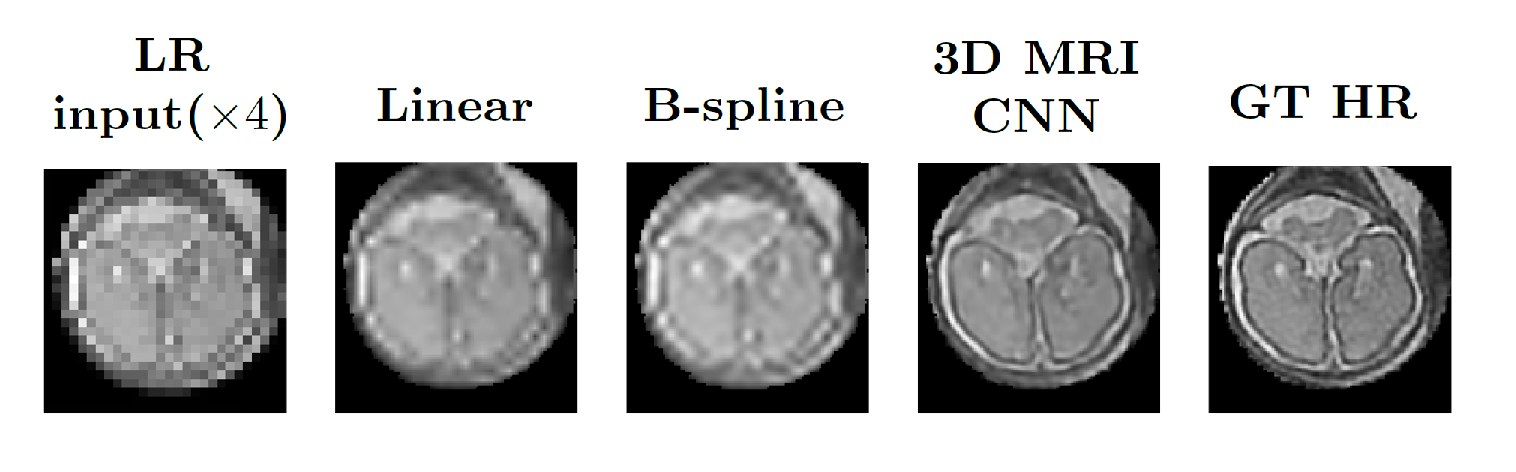}\label{}}\hfill
     \caption{SR applied to LR ($\times4$) volume reconstructed input. Benefits of learning the specific non-linearities to recover sharp edge gradients and improved high frequency signal components of the modality become more evident \emph{cf.} baselines as the amount of information required to upsample-successfully increases.}
     \label{fig:recon_upsample}
\end{figure}

We additionally perform preliminary experiments towards integrating network-based SR components more tightly with an SVR pipeline by investigating the ability of the network to upsample LR voxel intensities that result from an initial volume reconstruction iteration. Successful integration of an iterative (learning-based) SR and volume reconstruction loop will facilitate the well understood mutual benefits of reduced-motion SR input and improved input fidelity for the motion correction task. Qualitative comparison of ($\times4$) LR volume-reconstructed input and resulting upsampled results are found in Fig.\ref{fig:recon_upsample}. The benefit of learning the upsampling with modality specific data can be observed to manifest as sharper edge gradients and improved high frequency signal components. The visual quality gap between the baselines and our method can be seen to widen as the prior information required to successfully upsample at larger factors make the task more challenging.

\begin{figure}[!h]
\vspace{-7.5mm}
	\centering
	\resizebox{0.7\textwidth}{!}{
	\begin{tabularx}{\linewidth}{@{}cXX@{}}
		\multirow{-4}[2.5]{*}{
		\subfloat[]{%
			\includegraphics[width=0.17\columnwidth,height=0.17\columnwidth,angle=180, trim= 20 20 20 20,clip]{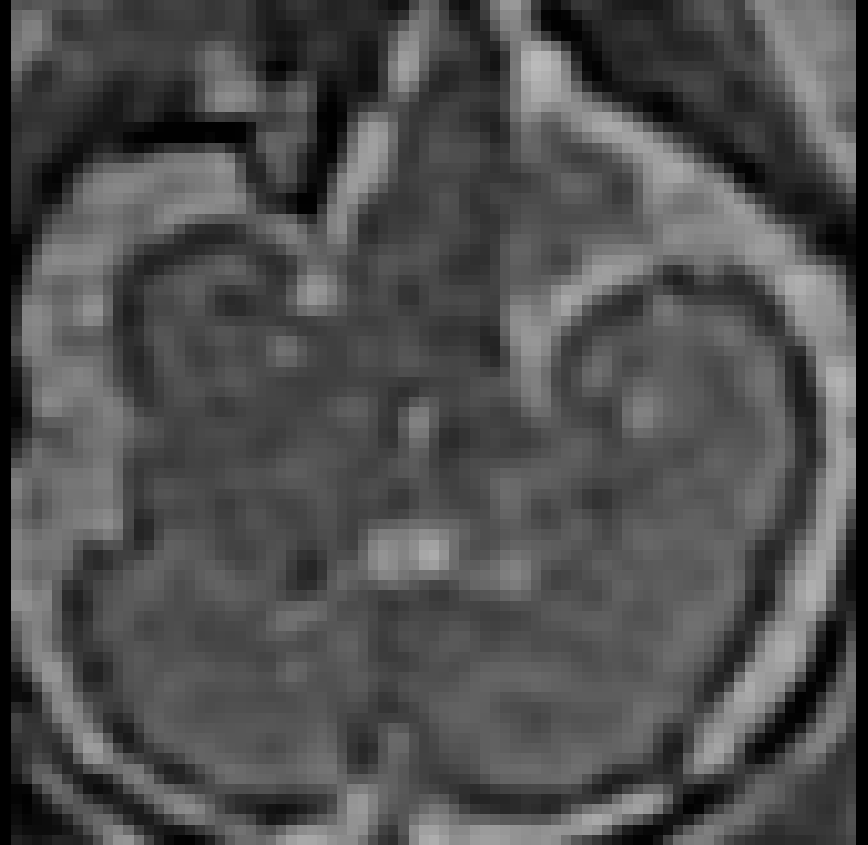}}
			}
		&
		\begin{tabular}{cccc}
			\vspace{-2pt}\subfloat[]{%
			\includegraphics[width=0.17\columnwidth,height=0.17\columnwidth,angle=180, trim= 20 20 20 20,clip]{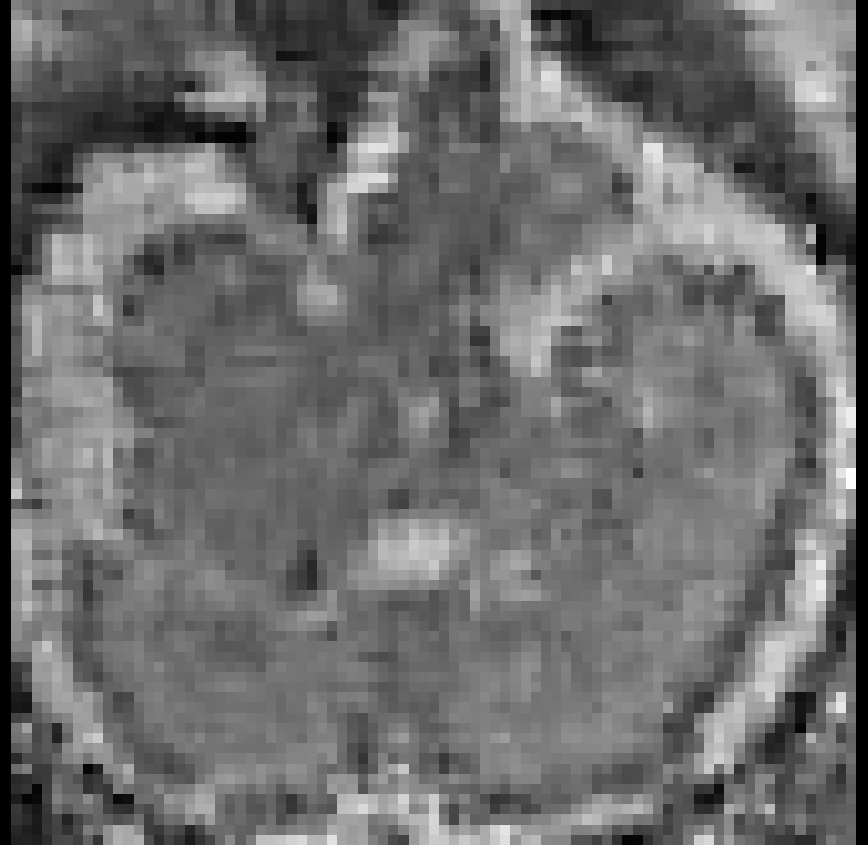}}	
			&	
			\vspace{-2pt}\subfloat[]{%
			\includegraphics[width=0.17\columnwidth,height=0.17\columnwidth,angle=180, trim= 20 20 20 20,clip]{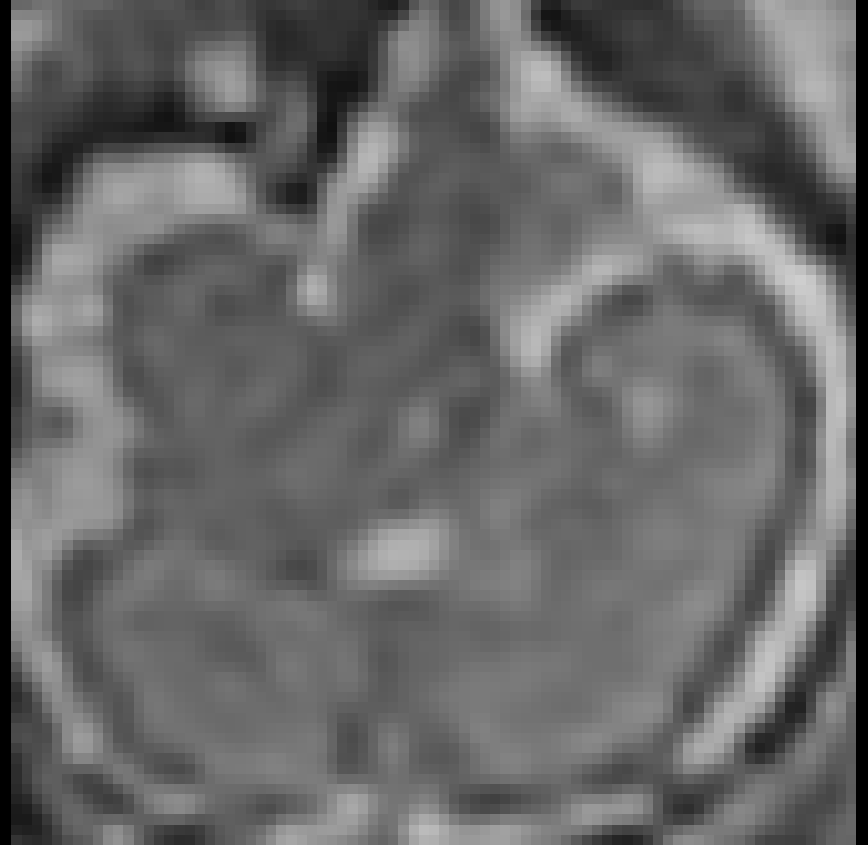}}
			&
			\vspace{-2pt}\subfloat[]{%
			\includegraphics[width=0.17\columnwidth,height=0.17\columnwidth,angle=180, trim= 20 20 20 20,clip]{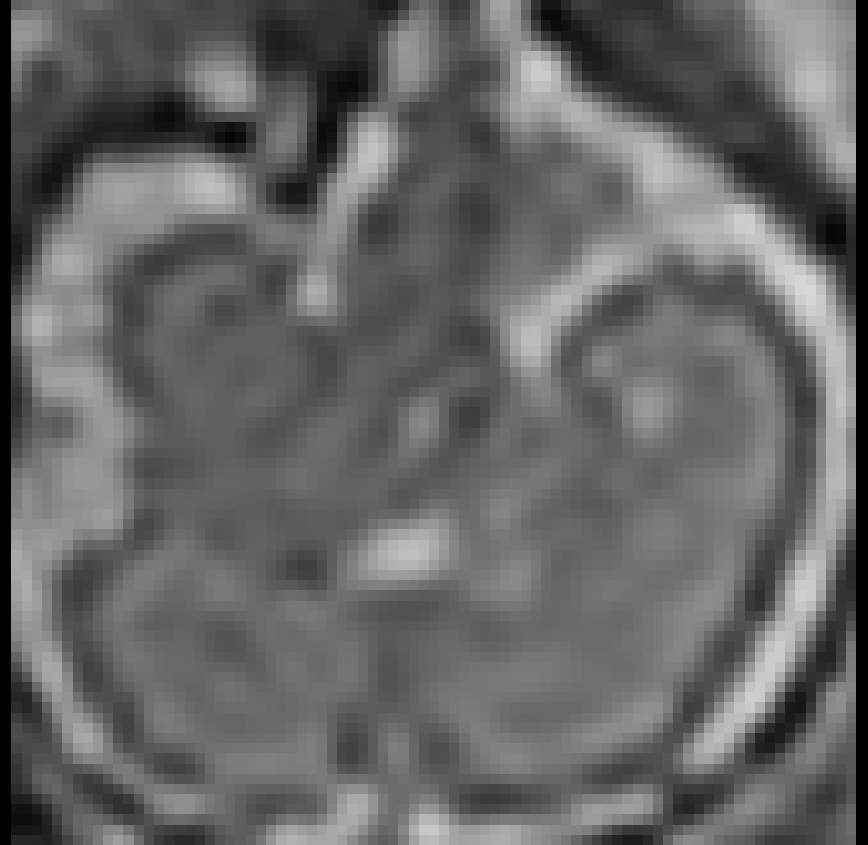}}	
			&
			\vspace{-2pt}\subfloat[]{%
			\includegraphics[width=0.17\columnwidth,height=0.17\columnwidth,angle=180, trim= 20 20 20 20,clip]{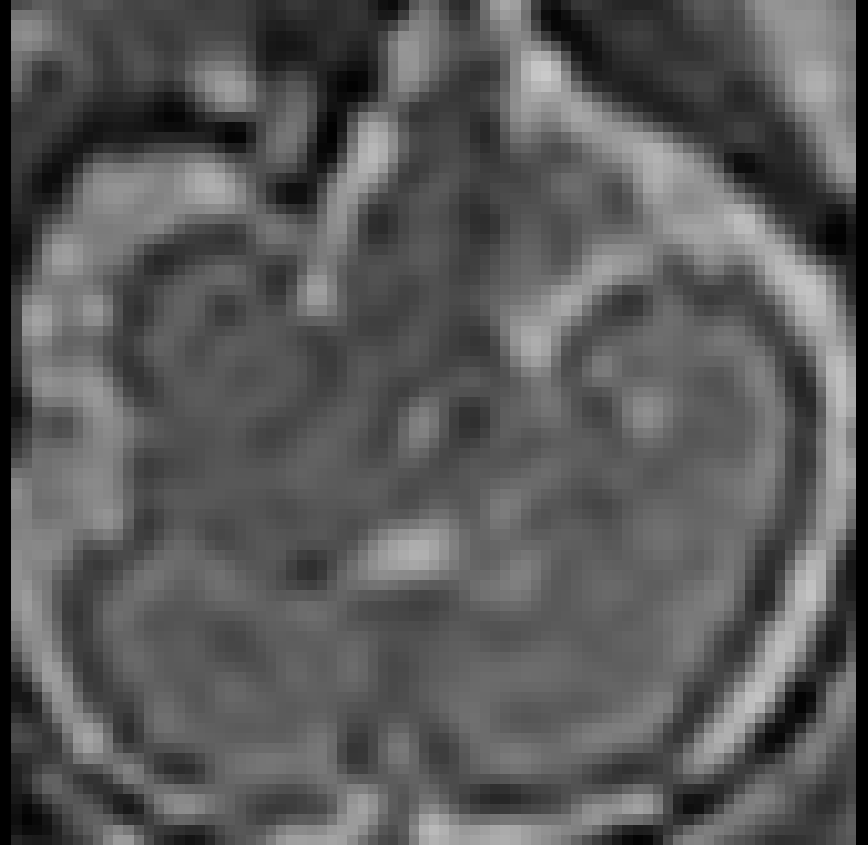}}
			\\
			\vspace{-2pt}\subfloat[]{%
			\includegraphics[width=0.17\columnwidth,height=0.17\columnwidth,angle=180, trim= 20 20 20 20,clip]{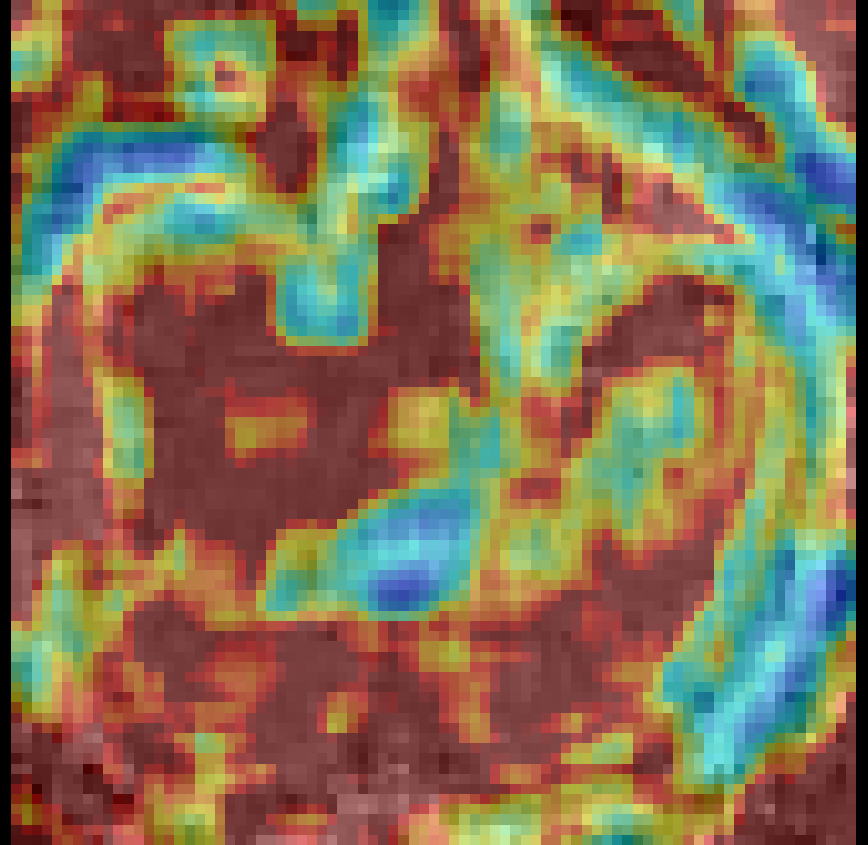}}	
			&
			\vspace{-2pt}\subfloat[]{%
			\includegraphics[width=0.17\columnwidth,height=0.17\columnwidth,angle=180, trim= 20 20 20 20,clip]{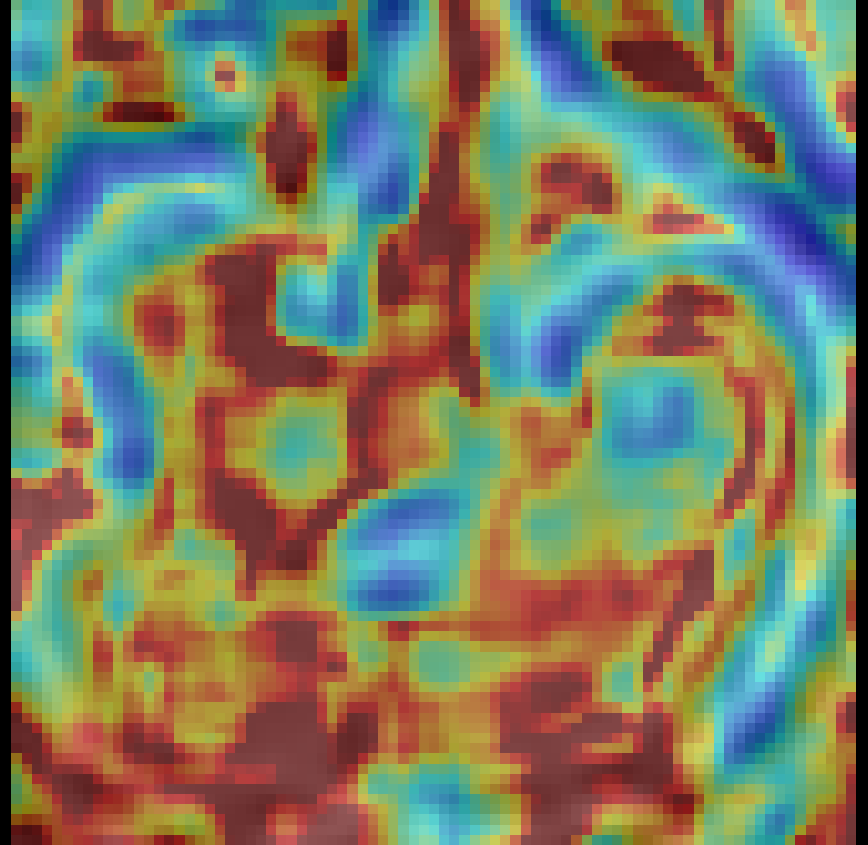}}
			&
			\vspace{-2pt}\subfloat[]{%
			\includegraphics[width=0.17\columnwidth,height=0.17\columnwidth,angle=180, trim= 20 20 20 20,clip]{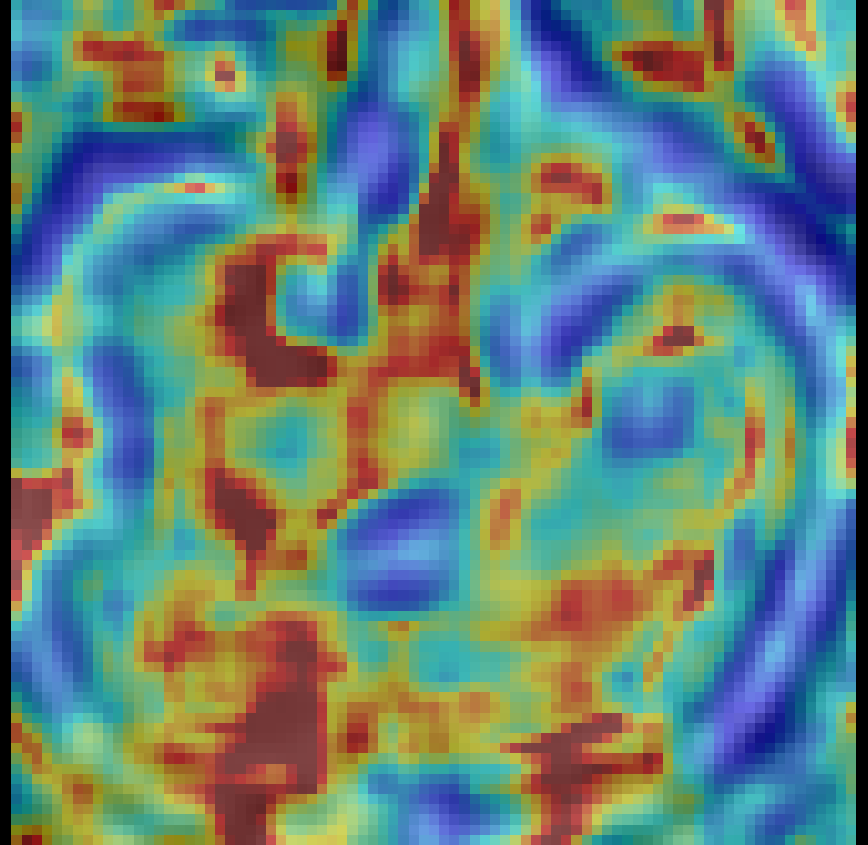}}	
			&
			\vspace{-2pt}\subfloat[]{%
			\includegraphics[width=0.17\columnwidth,height=0.17\columnwidth,angle=180, trim= 20 20 20 20,clip]{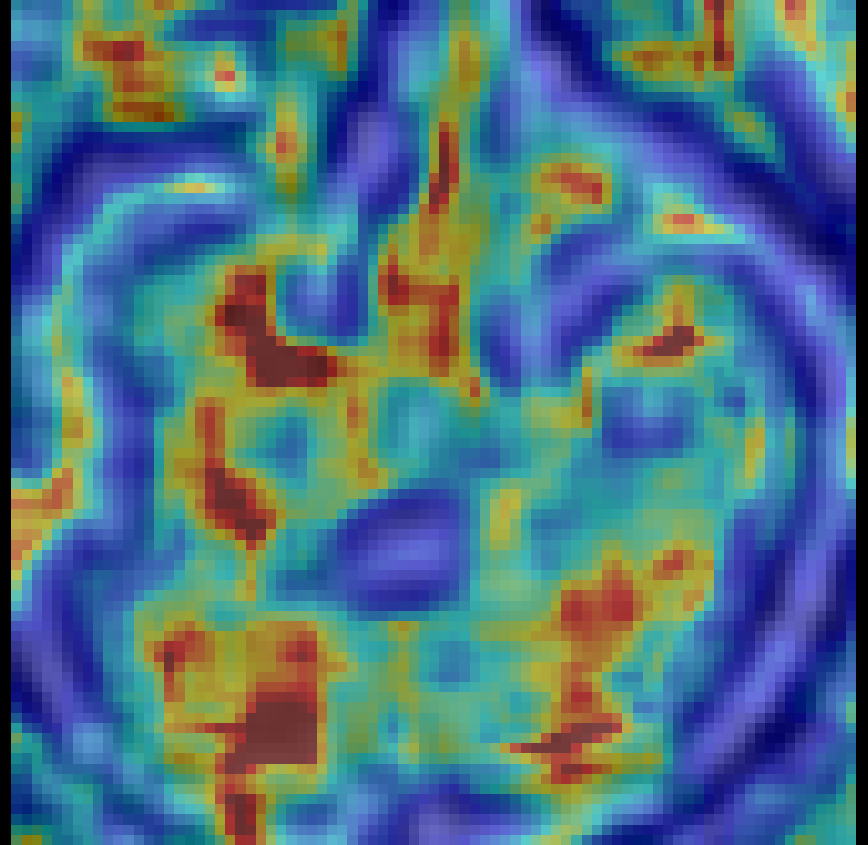}}
			\end{tabular}
	\end{tabularx}
}

\caption{(a) 2D slice through a fetal brain reconstruction, resulting from HR input-imagery. Attempting similar reconstruction from faster to acquire LR imagery, at half the in-plane resolution, results in highly degraded visual reconstruction quality (b) and gross DSSIM disparity (\emph{ie}. red heatmap regions) (f) with respect to the HR reconstruction. Na\"ive up-sampling $(\times2)$ of the LR in-plane input prior to reconstruction, with linear interpolation or B-splines, result in over-smoothed input. Loss of sharp gradient information and input-image fidelity can be seen to propagate to the respective reconstructions (c), (d) and disparity, with regard to the HR reconstruction, remains high (g), (h). Our 3D MRI CNN upsampling affords input closer to the original HR imagery and results in improved reconstructions (e) and reduced DSSIM (i) with visibly cooler heatmap regions (standard \emph{jet} color scale).}
\label{fig:dssim}
\end{figure}

\noindent \textbf{Volume Reconstruction Improvement:} 
In our third experiment we evaluate SVR performance using LR input stacks, upsampled by the considered strategies, before initiating the volume reconstruction task. We additionally perform SVR reconstruction with original HR imagery to provide the ``ground-truth'' reference brain volumes. Employing the three quality metrics, introduced previously, we evaluate how well super-resolved LR stack reconstructions correspond to the reconstructions due to original high, in-plane, resolution imagery. Table \ref{tab:psnr_ssim_cc} reports PSNR, SSIM and cross-correlation metrics for volume comparison (SR strategy with respect to ``ground-truth'' volume) for the $13$ patients that define the MRI stack test set. Super-resolving the LR input data with the proposed learning based approach can be observed to facilitate reconstruction improvement, across the investigated metrics. Visual evidence supporting this claim is found in Fig.~\ref{fig:dssim} (best viewed in color). Fig.~\ref{fig:dssim} displays 2D slices of patient fetal brain reconstructions resulting from the original HR input-imagery (far left) and identically spatially-located slices (a) resulting from (b) LR imagery (half the in-plane resolution), (c-d) input using na\"ive up-sampling strategies and (e) our 3D MRI CNN upsampling. Corresponding Structural Dissimilarity (DSSIM) error heatmaps (second row) provide improved visual spatial congruence between HR ground-truth and our method, supporting the claim that utilizing sensor specific priors is of marked benefit for the task of MRI fetal brain reconstruction from LR imagery.

\begin{table}[!h]
\begin{center}
\resizebox{0.7\textwidth}{!}{
	\begin{tabular}{*4c}    
		\toprule
		{Upsample} & PSNR\emph{[dB]} & SSIM & Cross-correlation  \\
		\midrule
		\rowcolor{white!50} ~\emph{No upsample}             ~~& $18.466 \; \Mypm 1.88$            ~~&~~ $0.534 \, \Mypm 0.15$          ~~&~~ $0.699 \, \Mypm 0.12$ \\ 
		\rowcolor{white!50} ~Linear              ~~& $19.268 \; \Mypm 1.14$            ~~&~~ $0.665 \, \Mypm 0.08$          ~~&~~ $0.815 \, \Mypm 0.06$ \\
		\rowcolor{white!50} ~B-Spline            ~~& $19.985 \; \Mypm 1.52$            ~~&~~ $0.698 \, \Mypm 0.12$          ~~&~~ $0.836 \, \Mypm 0.08$ \\
		\rowcolor{white!50} ~3D MRI CNN  ~~& $\mathbf{21.715 \, \Mypm 1.84}$   ~~&~~ $\mathbf{0.779 \, \Mypm 0.10}$ ~~&~~ $\mathbf{0.885 \, \Mypm 0.07}$ \\
		\bottomrule
		\hline
	\end{tabular}
}
\end{center} 
\caption{PSNR, SSIM and Cross-correlation evaluating disparity between reconstructed volumes using upsampled LR input (Linear, B-Spline, 3D MRI CNN) and ground-truth volumes.}
\vspace{-10mm}
\label{tab:psnr_ssim_cc}
\end{table}

\label{sec:conclusion}
\vspace{-2mm}
\section{Discussion and Conclusion}
\vspace{-1mm}
We introduce a 3D MRI CNN to upsample low resolution MR data prior to performing volumetric motion compensation and SVR reconstruction.
Our method produces upsampled images and uses them to reconstruct volumetric fetal brain representations that quantitatively outperform on reconstruction tasks that utilise conventional upscaling methods. This contribution helps to address the well-understood image resolution challenge in fetal brain MRI. Analysis of accuracy metrics, assessing upsampling quality, exhibit a mean PSNR increase of $1.25$ $dB$. Furthermore, when utilizing the upsampled imagery as SVR input, reconstructed fetal brain volumes show improvements of up to $1.73$ $dB$ over the provided baseline. In addition to quality improvement, 3D MRI CNN upsampling provides a computationally efficient approach affording an ability to initially image at lower resolutions, with a shorter acquisition time, thus provides faster and safer scanning for high-risk patients like pregnant women.

The current work has implicitly provided evidence that the method learns the PSF of the investigated MRI data well. In future it would be valuable to investigate this further, explicitly. Real-world LR/HR samples, acquired from scanners at differing resolutions, would allow quantitative evaluation of the ability to reconstruct physical scanner PSF and would further allow investigation of a model's ability to generalise to the reconstruction of PSFs not explicitly seen at training time. Further to this; the current work only investigates a single problem instance under one image modality. Future work will look to investigate the generalisability of the proposed framework to additional problem domains.



\bibliographystyle{splncs03}
\bibliography{references3}

\end{document}